\documentclass[]{article}
\usepackage{proceed2e}

\usepackage{times}
\usepackage{helvet}
\usepackage{color}
\usepackage{amssymb, amsmath}
\usepackage{graphicx}
\usepackage{enumitem}
\usepackage{subcaption}
\usepackage{multirow}
\usepackage{array}
\usepackage{natbib}
\usepackage{mathtools}
\usepackage[ruled]{algorithm2e}
\usepackage[absolute]{textpos}
\usepackage{hyperref}

\title{Evaluating Causal Models by Comparing Interventional Distributions}

 \author{
 Dan Garant\\
 dgarant@cs.umass.edu\\
 College of Information and Computer Sciences\\
 University of Massachusetts Amherst
 \And
 David Jensen\\jensen@cs.umass.edu\\
 College of Information and Computer Sciences\\
 University of Massachusetts Amherst
 }
\begin{document}
    
\maketitle
    
\begin{abstract}
The predominant method for evaluating the quality of causal models is to measure the graphical accuracy of the learned model structure. We present an alternative method for evaluating causal models that directly measures the accuracy of estimated interventional distributions.  We contrast such distributional measures with structural measures, such as structural Hamming distance and structural intervention distance, showing that structural measures often correspond poorly to the accuracy of estimated interventional distributions. We use a number of real and synthetic datasets to illustrate various scenarios in which structural measures provide misleading results with respect to algorithm selection and parameter tuning, and we recommend that distributional measures become the new standard for evaluating causal models.
\end{abstract}

\section{INTRODUCTION}

Causal inference is grounded in estimation of interventional effects. This requires researchers to identify \emph{which} variables are going to change as the result of an intervention and \emph{in what way} those variables can be expected to change. The former question, regarding \emph{which} variables are affected by a manipulation, is a structural question that requires correct identification of the causes of each variable. The latter question requires correctly representing the functional relationships between variables in the model.

\begin{figure}
\captionsetup[subfigure]{labelformat=empty}
\centering
\begin{subfigure}{0.3\linewidth}
\includegraphics[width=\linewidth]{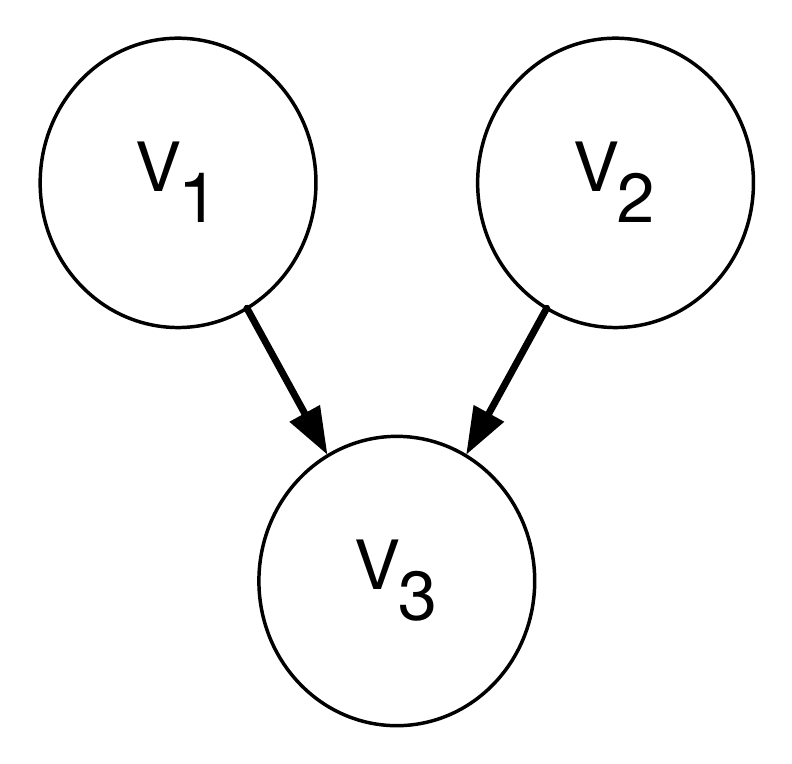}
\caption{$G_1$}
\label{fig:three-var-example-full}
\end{subfigure}
\begin{subfigure}{0.3\linewidth}
\includegraphics[width=\linewidth]{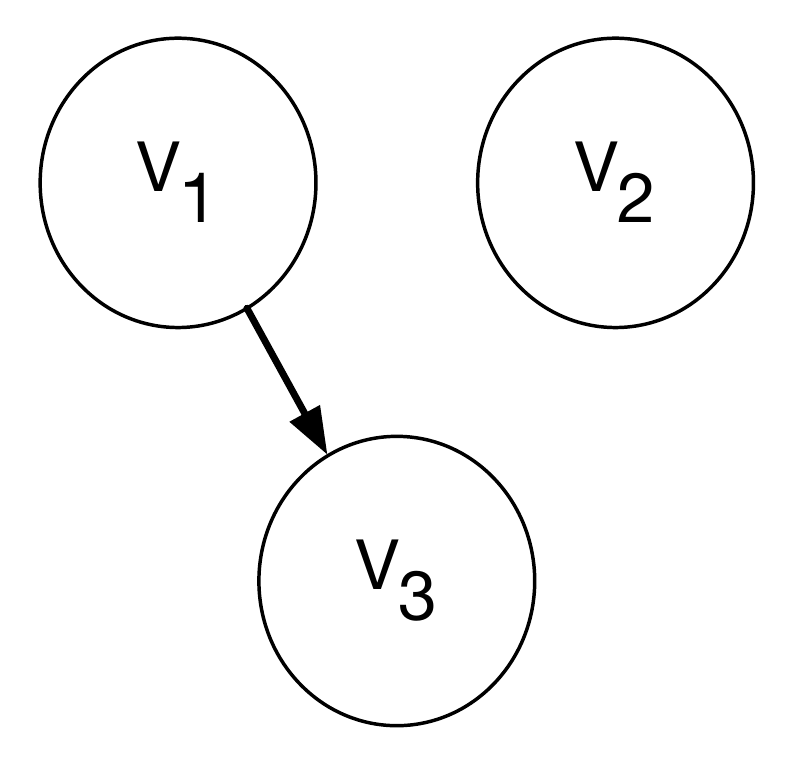}
\caption{$G_2$}
\label{fig:three-var-example-omit-2}
\end{subfigure}
\begin{subfigure}{0.3\linewidth}
\includegraphics[width=\linewidth]{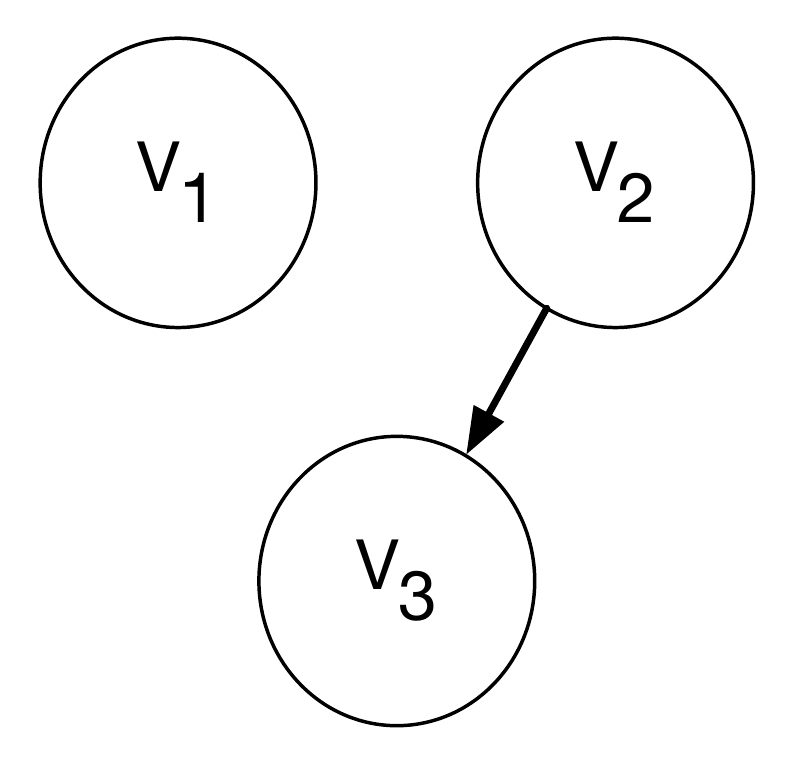}
\caption{$G_3$}
\label{fig:three-var-example-omit-1}
\end{subfigure}
\caption{Variants of a 3-variable system. For $G_1$, $V_3 \!\sim\!\mathcal{N}(V_1\!+\!0.1V_2, 1)$. In $G_2$, $V_3\!\sim\!\mathcal{N}(V_1, 1)$. In $G_3$, $V_3\!\sim\!\mathcal{N}(0.1V_2, 1)$. In all cases, $V_1, V_2 \sim \mathcal{N}(0, 1)$.}
\label{fig:three-var-example}
\end{figure}

The prevailing approaches to causal discovery from observational data focus on identifying the correct causal structure, represented as a directed acyclic graph (DAG). DAG models are almost always evaluated using graph-based measures of quality, such as structural Hamming distance (SHD)~\citep{tsamardinos2006maxmin} and structural intervention distance (SID)~\citep{peters2015structural}.
These structural quantities measure the quality of an estimated DAG by comparing the estimated edge set to a known edge set.
Such measures only characterize part of the causal inference task, specifically, \emph{which} variables are affected by a potential manipulation.
However, in many settings, the ultimate quantity of interest is the \emph{interventional distribution},
which completely characterizes the nature of a causal relationship.
As we will show, SHD and SID can be poor proxies for the quality of estimated causal effects or interventional distributions. In particular, SHD often overestimates the consequences of model over-specification (including too many edges), while SID imposes no penalty for over-specification. Conversely, under-specified models (with too few edges) can be problematic, but will impact distributional quality in a manner that is consistent with the strength with which omitted variables affect others in the model.
For example, consider the models shown in Figure~\ref{fig:three-var-example}.  With respect to the true graph $G_1$, $G_2$ and $G_3$ both have SHD and SID of 1, but omission of $V_1$ induces more severe parameterization errors than omission of $V_2$. The consequences of model over-specification are dependent in part on estimator selection and sample size, but SID and SHD do not account for such factors.

We present an evaluation methodology for observational causal discovery techniques.
This methodology relies on distributional distances for evaluation of an estimated parameterized DAG, and takes both structural quality and parametric quality into account.
We demonstrate several desirable properties of distributional distances, and show how these distances provide a more complete and accurate characterization of various modeling errors than commonly employed structural error measures. 
To identify the practical differences between distributional distances and structural distances, we performed exhaustive experimentation on three real domains and a number of synthetic domains commonly used in the literature. We highlight a number of instances in which structural distances can mislead researchers aiming to compare algorithms for learning and inference with causal models.

\section{CAUSAL GRAPHICAL MODELS}

Causal graphical models, represented as parametrized directed acyclic graphs, are an attractive framework for estimating interventional distributions from observational data. A DAG $G$ has a set of vertices, $\mathbf{V}(G)$, and a set of edges $\mathbf{E}(G)$. Each vertex $v$ has an associated conditional probability model $P(v|\mathbf{PA}^G_v=\vec{p})$, specified by the vector of values $\vec{p}$ of $v$'s parent variables $\mathbf{PA}^G_v$. When clear from context, we will write this parent set as $\mathbf{PA}_v$.

Estimating causal quantities with DAGs is simplified through a number of graphical procedures and criterion.
The \emph{do}-Calculus \citep{galles1995testing} specifies a graphical procedure for testing \emph{identifiability} of a causal effect, estimating the effects of an intervention given a known parameterized DAG. 
A causal effect is said to be \emph{identifiable} if it can be estimated from observed quantities.
The associated \emph{do} operator is a notational convenience to indicate that a probabilistic expression is related to a specific interventional context. This machinery is necessary because, in the general setting, an interventional distribution, e.g., $P(O|do(T=t_1))$, is distinct from an observational conditional distribution, $P(O|T=t_1)$. The former specifies a probability distribution over $O$ where $T$ is forced to take on the value $t_1$, whereas the latter specifies a distribution over $O$ where $T$ is observed as $t_1$. When intent is clear, we will abbreviate $P(O|do(T=t_1))$ as $P(O|do(t_1))$.

A nearly universal characteristic of observational data is the presence of \emph{back-door paths} between a treatment $T$ and an outcome $O$ of interest. A non-directed path $T v_1 v_2 v_3 \ldots v_n O$ in a DAG $G$ is called a \emph{back-door path} when $v_1$ is a parent of $T$. Following the rules of \emph{d}-separation \citep{pearl2009causality}, a set of variables $\mathbf{Z}$ that blocks every back-door path from $T$ to $O$, such that no member of $\mathbf{Z}$ is a descendant of $T$, is said to be a valid back-door adjustment set for $(T, O)$. In this case, $\mathbf{Z}$ is said to satisfy the \emph{back-door criterion} and $P(O|do(T=t)) = \sum_{\vec{z}} P(O|T=t, \mathbf{Z} = \vec{z})P(\mathbf{Z} = \vec{z})$ \citep{pearl2009causality}. \cite{shpitser10validity} presented a  relaxation of the back-door criterion for which $\mathbf{Z}$ permits identification of causal effects between $T$ and $O$, often referred to as the \emph{generalized back-door criterion}.

\section{EXISTING EVALUATION METHODS}
\label{sec:existing-evaluation-methods}

Structural Hamming distance (SHD) \citep{tsamardinos2006maxmin,acid2003searching} is commonly used to measure of distance between DAGs \citep{de2009comparison,kalisch2007estimating,pellet2008using,hoyer2009nonlinear,colombo2012learning,hyttinen2014constraint}.\footnote{SHD is sometimes decomposed into true/false positive rates, or the number of missing/extra/incorrectly oriented edges} SHD measures the number of edge additions, deletions, or reversals necessary to transform one DAG into another.
SHD has become a common measure for evaluating causal discovery algorithms. However, as shown by \cite{peters2015structural}, a non-zero SHD is not necessary for consistent estimation of causal effects. A simple example can be gleaned from Figure~\ref{fig:three-var-example}, treating $G_3$ as the true causal structure and $G_1$ as the estimated structure. In this case, $\mathit{SHD}(G_3, G_1) = 1$, but all interventional distributions are consistently estimated \citep{galles1995testing}. 

\cite{peters2015structural} proposed a measure of structural quality, called \emph{structural intervention distance} (SID), that counts the number of interventional distributions that are inconsistently estimated by a model. Specifically, with respect to a true DAG $G$ and an estimated DAG $H$, SID is computed as the number of pairs of variables $(V_1, V_2)$ for which:
\begin{itemize}[topsep=0pt,parsep=0pt]
    \item $V_1 \in \mathbf{PA}^G_{V_2}$ and $V_2 \in \mathbf{PA}^H_{V_1}$, or
    \item $\mathbf{PA}^{H}_{V_1}$ is not a valid adjustment set for $P(V_2|do(V_1))$.
\end{itemize}

Thus, a SID of zero is necessary for consistent estimation of causal effects.
However, SID is insensitive to model over-specification, since any set of variables $\mathbf{Z}$ such that $\mathbf{PA}^{G}_{V_2} \subseteq \mathbf{Z} $ is a valid adjustment set for $P(V_1|do(V_2))$. Thus, when a DAG model $H$ is a super-graph of the true graph $G$, $\mathit{SID}(G, H) = 0$. Over-specification permits consistent estimation in the large-sample limit, but dense models can dramatically reduce statistical efficiency \citep{koller2009probabilistic}. Unlike SID, SHD does penalize for model over-specification, but with equal weight as model under-specification.
\citeauthor{peters2015structural} proposed a modification of SID that penalizes superfluous edges by counting the difference in the number of edges between $G$ and $H$. As with SHD, this penalization supposes that all edges are equally important.

\section{DISTRIBUTIONAL DISTANCES} 

In many real situations, directed acyclic graphs are not the ultimate artifact of interest---they are a representation that facilitates estimation of interventional effects \citep{pearl2009causality,spirtes2000causation}. Thus, it seems natural to define an accuracy measure in terms of interventional effects rather than graphical structure. Most causal quantities of interest take the form of probability queries with $do$ operators, for instance $P(O|do(T=1))$. These quantities can be estimated by a learned distribution $\hat{P}$ using a parameterized DAG or another causal modeling technique. The accuracy of the $O-T$ interventional distribution can be assessed by comparing the true distribution $P$ to the estimated distribution $\hat{P}$ using an information-theoretic metric.

Despite the simplicity of this formulation, few researchers evaluate their models using direct comparison of known distributions to estimated distributions. Notable exceptions are \cite{tsamardinos2006maxmin} and \cite{eaton2007bayesian}. However, neither of these works consider the intrinsically causal task of interventional distribution estimation. \citeauthor{tsamardinos2006maxmin} use an information theoretic measure to compare estimated \emph{predictive} distributions to true \emph{predictive} distributions. In this work, we explore the use of total variation distance (TV) \citep{lin1991divergence} to measure distance between two \emph{interventional} distributions for an outcome $O$. For discrete outcomes, this computation is quite straightforward:
\begin{align}
\label{eq:total-variation}
TV_{P, \hat{P}, T=t}(O) = \frac{1}{2} \sum_{o \in \Omega(O) } \big| &P \left( O = o|do(T=t) \right) - \notag\\ &\hat{P}\left(O=o|do(T=t) \right) \big|,
\end{align}
where $\Omega(O)$ is the domain of $O$.
For continuous distributions, TV can be computed through an integral of differences in probability densities.
Total variation characterizes both the parametric quality and the structural quality of a model. When a model is over-specified, statistical efficiency degrades and the estimator $\hat{P}$ will have high variance. When a model is under-specified, $\hat{P}$ may not be a consistent estimator of $P$. TV has the advantage of penalizing model errors in accordance with their impact on the quality of probability estimates, rather than treating all errors as having equal weight as in SHD or SID. Although TV is not constrained to application on a DAG, we can summarize the quality of an estimated DAG $\hat{G}$ by computing a sum of pairwise total variations:
\begin{align}
\label{eq:sum-pairwise-variations}
TV_{\mathit{DAG}}(G, \hat{G}) = \sum_{\mathclap{V \in \mathbf{V}(G), V' \in \mathbf{V}(G) \setminus \{V\}}} TV_{P_{G}, P_{\hat{G}}, v'=v'_*}(V) 
\end{align}
Here, $v'_*$ represents the value of the hypothetical intervention to $V'$ and is a fixed value assigned by the analyst. For instance,
$v'_*$ could be set to a large value based on the quantiles of $V'$. More generally, we could consider a sum or an integral over settings of $v'_*$, however it seems unlikely that this added expense would yield more informative results.

Evaluating $TV_{\mathit{DAG}}$ requires inference, which may be computationally expensive. It is common to have a clearly defined set of treatments and outcomes of interest, in which case the sum of pairwise total variations would be best expressed in terms of only those vertices. If an investigator truly is interested in all pairwise interventional distributions, then evaluating this sum is no more expensive than \emph{using} the model to reason about causal effects.

There is a clear relationship between structural intervention distance and $TV_{\mathit{DAG}}$. 
In particular, when $TV_{\mathit{DAG}}$ is 0, then for all $V$ and $V'$, $P_{\hat{G}}(V|do(V'=v'_*)) = P_{G}(V|do(V'=v'_*))$. 
As defined by~\cite{peters2015structural}, SID counts the number of pairs $(V, V')$ such that $P_{\hat{G}}(V|do(V'=v'_*)) \neq P_{G}(V|do(V'=v'_*))$, thus $TV_{\mathit{DAG}}(G, \hat{G}) = 0 \Rightarrow \mathit{SID}(G, \hat{G}) = 0$. However, the converse of this statement does not hold. 
$TV_{\mathit{DAG}}(G, \hat{G})$ depends in part on how well the parameters of $\hat{G}$ have been estimated. 
It is possible that $\hat{G}$ permits unbiased inference, that is, the parent set of each node $V$ is a valid adjustment set for an intervention on $V'$, but due to variance in finite-sample settings, $P_{\hat{G}}$ does not exactly equal $P_G$. $TV_{\mathit{DAG}}$ accounts for both the bias \emph{and} variance of the estimated interventional distribution, and is therefore more closely related to real-world use cases for causal discovery.

\subsection{A SIMPLE EXAMPLE}

Before examining how structural distances compare to total variation distance, consider again the example presented in Figure~\ref{fig:three-var-example}.  If $G_1$ is the true model, then omission of the edge $V_2 \to V_3$ in $G_2$ results in an SHD of 1 (edge edit distance 1) and an SID of 1 (one interventional distribution is mis-estimated).  Similarly, omission of the edge $V_1 \to V_3$ in $G_3$ results in an SID and an SHD of 1. These two edge-omission errors are indistinguishable.

Now consider an information-theoretic evaluation. Consider the three alternate conditional models for $V_3$:
\begin{align}
    &P_1(V_3 | v_1, v_2) = \mathcal{N}(v_1 + 0.1v_2, 1) \\
    &P_2(V_3 | v_1, v_2) = P_2(V_3 | v_1) = \mathcal{N}(v_1, 1) \\
    &P_3(V_3 | v_1, v_2) = P_3(V_3 | v_2)  = \mathcal{N}(0.1v_2, 1)
\end{align}
$P_1$ corresponds to a correct model, $P_2$ represents an estimated model which omits $V_2 \to V_3$, and $P_3$ represents an estimated model which omits $V_1 \to V_3$. We computed $TV_{P_1, P_m, V_i=2}(V_3)$ for $m=2,3$ and $i=1,2$. We used adaptive quadrature to approximate integrals over $V_3$. Table~\ref{tbl:simple-example-tv} demonstrates that, consistent with the SID definition, $P_2$ and $P_3$ both mis-estimate one interventional distribution (there is one non-zero row per column). The key advantage of TV lies in its ability to differentiate between the \emph{severity} of the mis-estimation. In this case, omitting $V_1 \to V_3$ (TV=0.68) is a more significant error than omitting $V_2 \to V_3$ (TV=0.08). 

\begin{table}[ht]
\centering
\begin{tabular}{lrr}
  \hline
 & $TV_{P_1, P_2}(V_3)$ & $TV_{P_1, P_3}(V_3)$ \\ 
  \hline
$do(V_1 = 2)$ & 0.00 & 0.68 \\ 
$do(V_2 = 2)$ & 0.08 & 0.00 \\ 
   \hline
\end{tabular}
\caption{Total variation of $P_2$ and $P_3$ with respect to true model $P_1$ and hypothetical interventions on $V_1$ and $V_2$}
\label{tbl:simple-example-tv}
\end{table}

\section{EVALUATION METHODOLOGY}

To explore the differences between total variation distance and structural distances in realistic situations, we instrumented and gathered data from three real domains and three commonly used synthetic data generation techniques.

\begin{figure*}[ht!]
    \centering
    \includegraphics[width=\linewidth]{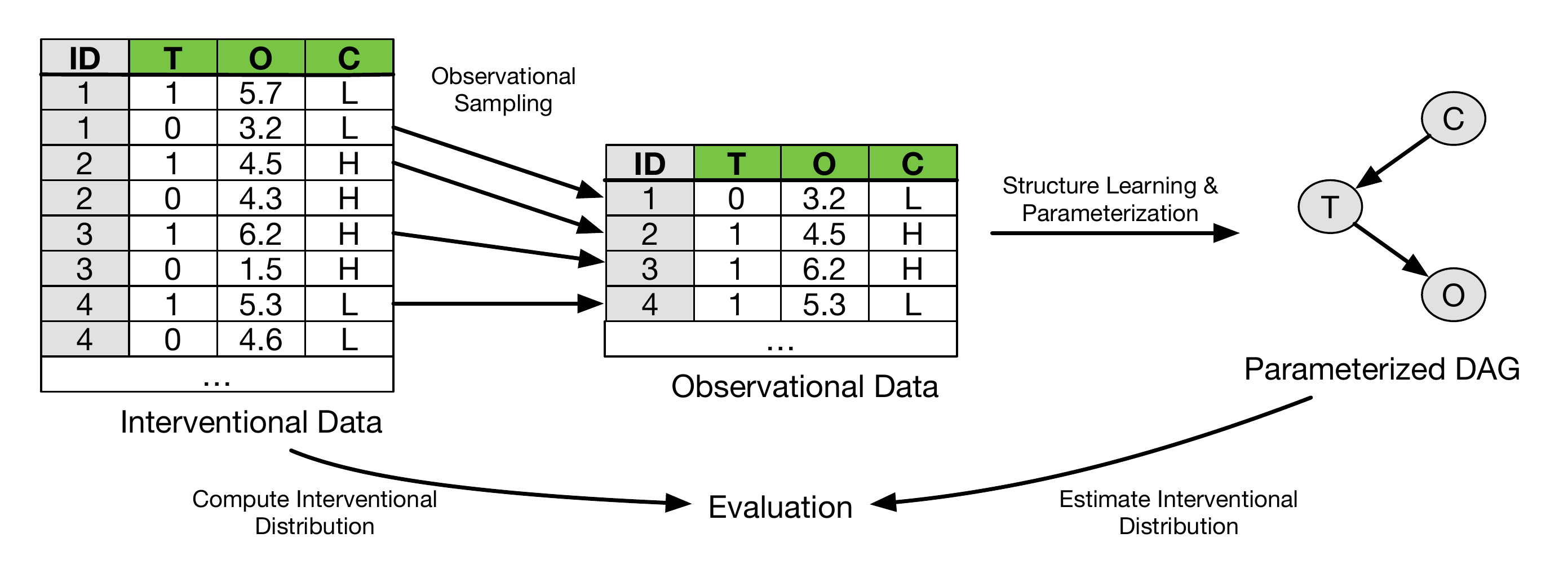}
    \caption{Outline of the evaluation process. An interventional distribution with multiple observations per subject (identified by \texttt{ID}) is sub-sampled using Algorithm~\ref{alg:passive-treatment} to form a smaller dataset containing one obervation per subject and observational bias. Causal discovery algorithms are employed to estimate a DAG from the observational data, and the resulting structure is parameterized with maximum likelihood estimation. Interventional distributions are estimated using the parameterized DAG, and compared to those implied by the interventional data.}
    \label{fig:evaluation-process}
\end{figure*}

\subsection{REAL DOMAINS}

Each real domain is a large-scale computational system, consisting of many thousands of lines of source code and used for a diverse set of tasks. Large-scale software systems offer several desirable characteristics for the purposes of empirical evaluation. Specifically, such systems are:

\begin{itemize}[leftmargin=5pt]
\item[] \textbf{Empirical:} They are pre-existing systems created by individuals other than the researchers for purposes other than evaluating algorithms for causal discovery.  This avoids implicit or explicit bias that can affect the structure and parameters of synthetic data generators.

\item[] \textbf{Stochastic:} They produce experimental results that are non-deterministic through some combination of epistemic factors (e.g., latent variables) and aleatory factors (inherently stochastic behavior in the data generating process). 

\item[] \textbf{Identifiable:} They are amenable to direct experimental investigation to estimate interventional distributions.  In particular, these systems facilitate interventions on single variables in ways that largely avoid the ``fat hand'' effects that plague some physical systems.

\item[] \textbf{Recoverable:} They lack memory or irreversible effects, which enables complete state recovery during experiments.  Such state recovery is far from simple because many modern software systems have features such as caches that can create temporal dependence among runs, but state recovery is possible in principle.  This enables factorial experiments in which every joint combination of interventions is run on every experimental unit.

\item[] \textbf{Efficient:} They are capable of generating large amounts of data that can be recorded with relatively little effort.

\item[] \textbf{Reproducible:} They allow future investigators to recreate nearly identical data sets with reasonable resources and without access to one-of-a-kind hardware or software.
\end{itemize}

Few, if any, other classes of systems offer a similar range and combination of advantages.

Within each computational system, we measure three classes of variables: subject covariates, treatment settings, and outcomes.
Outcomes are measurements of the result of a computational process.
Treatments correspond to system configurations and are selected such that they could plausibly induce changes in outcomes.
Subject covariates logically exist prior to treatment and are invariant with respect to treatment.
With these variables defined, we conduct a factorial experiment.
This dictates that each combination of treatment variables be applied to every subject.
Thus, given a set of $n$ subjects and $k$ binary treatment variables, there are $n 2^k$ data instances, referred to as \emph{subject-treatment combinations}. This dataset can then be used to estimate interventional distributions for the treatment variables in a straightforward manner without confounding bias.

\begin{algorithm}
\DontPrintSemicolon
\KwIn{Interventional dataset $I$, biasing strength $\beta \geq 0$, biasing covariate $C$}
\KwOut{Biased dataset $O$, $|O| = nd$}

$l \gets $ The number of distinct values of $C$ \;
\ForEach{Subject $e \in I$}
{
    Let $C_e \in \{1..l\}$ represent the $C$ value of subject $e$ \;
    $Assign \gets \{\}$ \;
    \ForEach{Treatment $T_j$}
    {
        $s_{ej} \gets \begin{cases} 
                1 & \text{if $C_e \times j$ is even} \\
                -1 & \text{if $C_e \times j$ is odd} \\
        \end{cases}$ \;
        $p \gets \text{logit}^{-1}(s_{ej} \beta)$\;
        $t_j \gets $ Bernoulli$(p)$ \;
        $Assign \gets Assign \cup \{ T_j = t_j \}$ \;
    }
    $M \gets $ Record in $I$ corresponding to $(e, Assign)$ \;
    $O \gets O \cup M$ \;
}

\caption{Logistic Sampling of Passive Treatments}
\label{alg:passive-treatment}
\end{algorithm}

We can also transform the dataset generated by the factorial experiment into a dataset that has properties consistent with observational data. This transformation induces a set of back-door paths between treatments and outcomes, yielding a dataset with a single treatment observation per subject.
Using a logistic function, Algorithm~\ref{alg:passive-treatment} samples a value for each treatment $T_j$ with strength of dependence $\beta$ and sign $s_{ej}$ depending on subject $e$'s value of $C$. For each domain, we note which variable acts as $C$, the biasing covariate. 

When $\beta$ is large ($\geq 3$), some subject-treatment combinations (with $s_{ej} = -1$) are almost always in the control setting ($P(T_j = 0 | C = C_e) \approx 0$), and some subject-treatment combinations (with $s_{ej} = 1$) are almost always in the treated setting ($P(T_j = 1 | C = C_e) \approx 1$). When $\beta$ is zero, the dataset corresponds to a uniformly randomized experiment. In this case, conditional distributions $P(O|T=t)$ yield consistent estimates of the causal quantities $P(O|do(t))$. For $\beta > 0$, back-door paths exist and this conditional model is no longer appropriate for causal reasoning, requiring causal learning and reasoning techniques appropriate for observational data.
Each domain is described below.
\footnote{Datasets are available at \url{https://kdl.cs.umass.edu/display/public/Causal+Evaluation}}

\subsubsection{Oracle Java Development Kit}
The Java Development Kit (JDK) is a software library used to compile, run, and diagnose problems with Java programs.
Each subject in this domain is an open-source Java project, and the computational process is compilation and execution of the unit tests for that project.
As treatments, we selected four system settings that are of interest to developers: compiler optimization, use of debugging symbols, garbage collection method, and code obfuscation.
For outcomes, we measured factors pertaining to the run time, memory usage, time to compile, and code size. To better approximate observational settings, we measured subject covariates which could confound treatments and outcomes if no controls were present. We measured the number of non-comment source statements in both the project source code and associated unit test source code, along with the number of functions and classes in the unit test source, correlating to some extent with unit test runtime.
The number of ``Javadoc'' comments in the unit test source code was also measured, as it may be associated with code quality---this was selected as the biasing covariate. 

\subsubsection{PostgreSQL}
PostgreSQL (or just Postgres) is a widely-used open-source database management system.
For this domain, a subject is a database query, and the computational task is to execute that query.
Treatments on the Postgres domain are system settings that a database administrator may be interested in tuning through experimentation: the use of indexing, page access cost estimates, and working memory allocation.
As outcomes, we recorded query runtime, the number of blocks read from shared memory, temporary memory, and a fast memory cache.
As subject covariates, we measured aspects of the query itself such as the number of joins, number of grouping operations, length, and statistics of the referenced tables. 
We also recorded the number of rows retrieved by the query (which is logically prior to treatment), using this as the treatment-biasing covariate.

\subsubsection{Hypertext Transfer Protocol}
The Hypertext Transfer Protocol (HTTP) is the primary mechanism by which information is transferred across the Web.
In this domain, a subject is a web request to a specific web site, and the computational process under study is the transmission of that request and the response it elicits.
We selected several options of the HTTP request as treatments: use of a proxy server, compression specifications, and the HTTP user agent.
Several response characteristics served as outcomes: the number of HTML attributes and tags, the elapsed time of the web request, the content length before decompression, and the size of the response after decompression.
Few treatment-invariant subject covariates exist in this domain, since almost every aspect of a web page is subject to change based on request parameters. The host-reported web server (e.g., Apache) is the sole subject covariate which is highly unlikely to be influenced by any of the above treatments; we used this as the biasing covariate. 

\subsection{SYNTHETIC DATA}

\subsubsection{Linear-Gaussian}
In our literature review, we found that synthetic linear-Gaussian systems were the most commonly used structures for evaluation   \citep{hyttinen2014constraint,ramsey2006adjacency,kalisch2007estimating,colombo2012learning}. The typical construction of such systems begins with generation of a random sparse DAG $G$ with an expected neighborhood size ($E[N]$) of 2, 3, or 5. The most common sparsity setting we found was $E[N] = 2$. Then, a weight matrix $\mathbf{W}$ is generated, with values sampled uniformly from $[0.1, 1]$ (the most commonly used interval in our review). A set of error terms $\vec{\epsilon}$ are generated from $\mathcal{N}(0, 1)$. Then, samples $X_i$ for each vertex $i$ are generated using the process:
\begin{align}
X_i \gets \sum_{j \in \mathbf{PA}_i^G} \mathbf{W}_{ji} X_j + \epsilon_i,
\end{align}

\subsubsection{Dirichlet}
Some authors have constructed synthetic DAGs using discrete variables with relationships arising from by a Dirichlet distribution. \cite{chickering2002finding} use a small number of gold-standard DAGs, for which the $E[N] \leq 2$. \cite{eaton2007bayesian} use the structure of the CHILD network \citep{cowell2007probabilistic}, which has $E[N] \approx 1$. Conditional probability tables are generated for $k$-state node $i$ using a Dirichlet distribution. Specifically, let $\mu = (\frac{1}{1}, \frac{1}{2}, \ldots, \frac{1}{k})$ and $\alpha = \frac{1}{\sum \mu}$. Number the joint assignments to $\mathbf{PA}_i$ as $1..A$. Consider rotations $\mu_a$ of $\mu$ such that $\mu_1 = (\frac{1}{k}, \frac{1}{1}, \ldots, \frac{1}{k-1})$ and $\mu_2 = (\frac{1}{k-1}, \frac{1}{k}, \ldots, \frac{1}{k-2})$. Then, for each numbered assignment $a$ to $\mathbf{PA}_i$, draw $P(X_i|\mathbf{PA}_i = {\mathbf{PA}_i}_a)$ from $\text{Dirichlet}(S\alpha\mu_a)$. The $S$ factor, often called the equivalent sample size, can be viewed as a measure of confidence in the Dirichlet hyper-parameters \citep{heckerman1995learning}. Unless otherwise noted, $S=10$ for our experiments, consistent with usage in the literature.

\subsubsection{Logistic}
A third category of synthetic data generation uses a logistic function to generate binary data. \cite{li2009controlling} largely follow the ``linear-Gaussian'' strategy. Random DAGs are generated with $E[N] \in \{2, 3\}$. Instead of weighting edges, the authors weight vertices, sampling $\vec{W}$ from $\{\delta, -\delta\}$. The strength of dependence parameter $\delta$ can be varied. In what follows, we use $\delta = 0.375$, the mean of the range explored in the original work. Then, values $X_i$ are sampled for vertex $i$ using:
\begin{align}
X_i \gets \text{Binomial}\left( \text{logit}^{-1} \left( \sum_{j \in \mathbf{PA}_i} X_j W_j \right) \right)
\end{align}
In some cases, it is useful to compare results on the synthetic datasets with those on our real datasets. In these cases, we use three ``look-alike'' configurations which each have the same number of variables and data points (subjects) as one of the real systems. In addition, the datasets agree on the number of treatments, outcomes, and subject covariates. The synthetic data generation strategy is adjusted to resemble a factorial experiment and the observational sampling process of Algorithm~\ref{alg:passive-treatment} is employed. This ensures that the synthetic domain and the real domain are comparable with respect to the back-door paths induced between treatments and outcomes. In what follows, we label these configurations with a J, P, or H depending on whether they were modeled after the JDK, Postgres, or HTTP domain, respectively. 

\subsection{ALGORITHMS}
We selected three algorithms representative of constraint-based, score-based, and hybrid causal discovery for our evaluation; respectively PC \citep{spirtes2000causation}, GES \citep{chickering2002finding}, and MMHC \citep{tsamardinos2006maxmin}. In PC and MMHC, we used the $G$-test with $\alpha=0.05$ for conditional independence testing on discrete data, and the $z$-statistic with Fisher's partial correlation for testing linear-Gaussian data. For GES and score-based phases of MMHC, we used BIC as the scoring criterion. In all cases, conditional probabilities were modeled with tables. These  choices are common, and are the default options in the \textsf{R} packages \texttt{pcalg} and \texttt{bnlearn} that we used in this study. 
PC and GES learn a complete partially directed acyclic graph (CPDAG), which may contain undirected edges as well as directed edges. CPDAGs represent a class of models with equivalent likelihood. When a CPDAG has more than one member, the reported performance value is a mean of the performance of each DAG extension. When a CPDAG has more than 100 DAG extensions, we sample 100 uniformly at random. We then measure SID, SHD, and TV on each DAG extension, and compute the mean of each measure for a given CPDAG.

\section{EMPIRICAL COMPARISON}

Ultimately, the distinction between total variation distance and commonly-used structural distance measures is important only if the two categories of evaluation would lead to different conclusions. We sought to address the following questions to help make that distinction clear:
\begin{itemize}[noitemsep,topsep=0pt,parsep=0pt,partopsep=0pt]
    \item Is the relative performance of causal discovery algorithms systematically different when evaluating with TV, as compared to a structural evaluation?
    \item Does model over-specification or under-specification impact total variation evaluations differently than evaluations with structural measures?
    \item Do the parameters used in synthetic data generation elicit different behaviors in TV than in SHD/SID?
\end{itemize}

\subsection{RELATIVE PERFORMANCE OF ALGORITHMS}

\begin{figure}
\centering
\begin{subfigure}{\linewidth}
    \includegraphics[width=\linewidth]{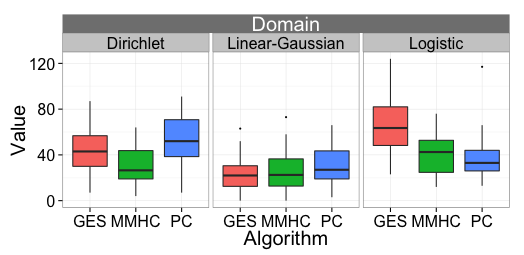}
    \caption{Structural intervention distance}
\end{subfigure}
\begin{subfigure}{\linewidth}
    \includegraphics[width=\linewidth]{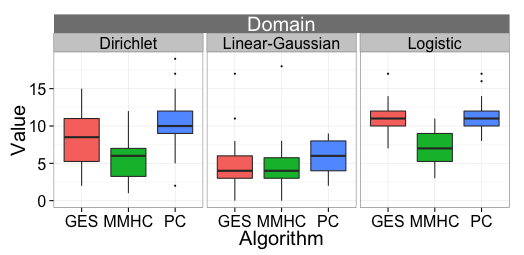}
    \caption{Structural Hamming distance}
\end{subfigure}
\begin{subfigure}{\linewidth}
    \includegraphics[width=\linewidth]{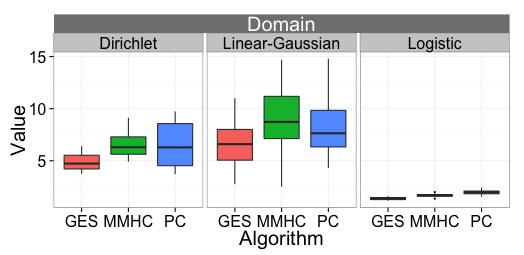}
    \caption{Total variation distance}
\end{subfigure}
\caption{Relative Performance on Synthetic Datasets}
\label{fig:relative-performance}
\end{figure}

One of the most basic questions about TV is whether it produces different conclusions than using SHD or SID in realistic evaluations.
We found that TV produces implies a very different ordering of the relative performance of different learning algorithms than that implied by SHD and SID.
We began by constructing 30 random DAGs with 14 variables and $E[N]=2$. 
We generated parameters on those DAGs using each of the synthetic data techniques and sampled 5,000 data points from each DAG. Then, we applied PC, MMHC, and GES to the resulting datasets and measured the SID, SHD, and sum of pairwise total variations as in equation~\ref{eq:sum-pairwise-variations}. As shown in Figure~\ref{fig:relative-performance}, some of the findings that would be reached with SID and SHD are not supported by a TV evaluation. The structural measures suggest that MMHC outperforms PC on the Dirichlet domain. However, the performance of the two algorithms is statistically indistinguishable as measured by TV. When measured with SID or SHD, GES does not outperform either MMHC or PC. However, GES is consistently the best performing algorithm in terms of interventional distribution accuracy.

\subsection{MODEL OVER-SPECIFICATION AND UNDER-SPECIFICATION}

\begin{table*}
\centering
{ \small
 \begin{tabular}{ lll | rrr | rrr | rrr  } 
  \hline
 Domain & Subjects & Model Type & \multicolumn{3}{c|}{SID: Min, Median, Max} & \multicolumn{3}{c|}{SHD: Min, Median, Max} & \multicolumn{3}{c}{TV: Min, Median, Max} \\ 
  \hline
\multirow{2}{*}{JDK} & \multirow{2}{*}{473} & Over-specify & 0 & \hspace{3em} 0 & 0 & 1 & \hspace{2.5em} 3 & 3 & 0.04 & \hspace{0.75em} 0.17 & 0.21 \\ 
  &  & Under-specify & 4 & 5 & 9 & 2 & 2 & 4 & 0.22 & 0.41 & 0.58 \\ 
  \hline
  \multirow{2}{*}{Postgres} & \multirow{2}{*}{5,000} & Over-specify & 0 & 0 & 0 & 0 & 1 & 2 & 0.00 & 0.06 & 0.09 \\ 
  &  & Under-specify & 4 & 6 & 8 & 3 & 4 & 5 & 0.17 & 0.35 & 0.61 \\ 
  \hline
  \multirow{2}{*}{HTTP} & \multirow{2}{*}{2,599} & Over-specify & 0 & 0 & 0 & 1 & 2 & 4 & 0.06 & 0.06 & 0.09 \\ 
  &  & Under-specify & 2 & 6 & 10 & 1 & 3 & 4 & 0.22 & 0.25 & 0.30 \\ 
   \hline
\end{tabular}
}
\caption{Evaluation Metric Comparison on Real Domains}
\label{tbl:altered-models}
\end{table*}

\begin{figure}
\centering
\includegraphics[width=\linewidth]{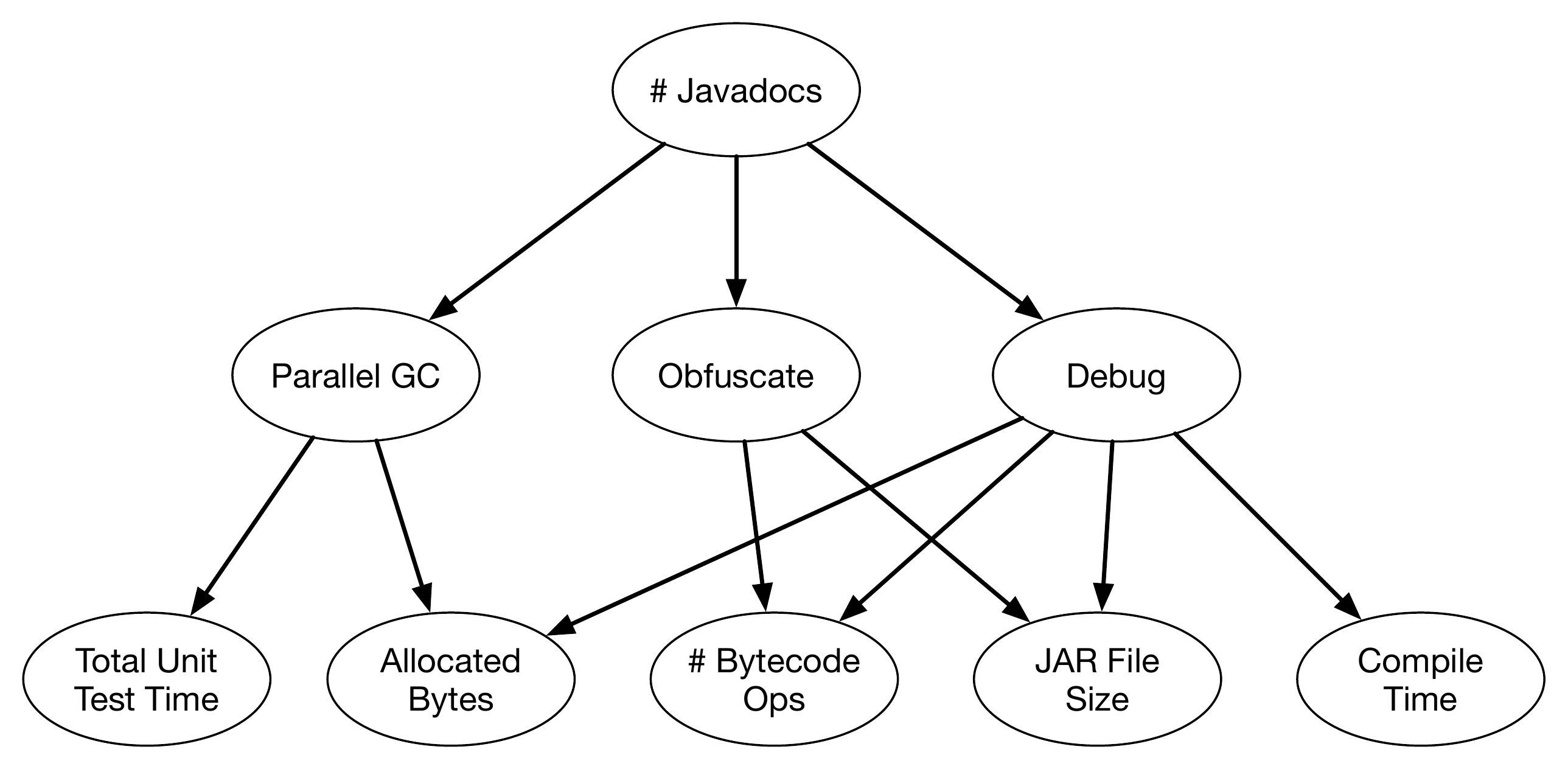}
\caption{Consistent Model for the JDK Dataset}
\label{fig:jdk-true-consistent-model}
\end{figure}

Another important question about TV is how the measure responds to specific types of errors in learned structure.  Specifically, we wanted to evaluate the effects of over-specification (extraneous edges) and under-specification (omitted edges) on model performance.  Compared to TV, we found that neither SID or SHD provide good proxies for the effects of over- and under-specification.
To characterize these effects, we turned to the real domains. In each case, treatment assignments are moderately biased ($\beta = 1$). For simplicity of illustration, we omit some subject covariates which we know cannot cause any of the treatments. From our experiments, we can identify which treatment-outcome pairs are causally related. We construct a true DAG by introducing an edge between each pair of causally related treatment and outcome. Since the biasing covariate necessarily blocks all back-door paths between each treatment and outcome, an edge is introduced between this covariate and all treatments. The resulting DAG model (illustrated for the JDK dataset in Figure~\ref{fig:jdk-true-consistent-model}) consistently estimates distributions $P(O|do(T=t))$ for all treatment-outcome pairs.

We altered the consistent models of each dataset to induce over-specification and under-specification. To quantify the effects of over-specification, we produced models in which one of the treatment variables had a directed edge into every outcome, regardless of the causal relationships in the true model. To quantify the effects of under-specification, we produced models in which one of the treatment variables had no outgoing edges. This process was then repeated for each of our three domains and each treatment variable within that domain. For each model, a sum of pairwise total variations were computed as $\sum_{T,O} TV_{P,\hat{P},T=1}(O)$, where $P$ represents the reference distribution given by the consistent model (as in Figure~\ref{fig:jdk-true-consistent-model}) and $\hat{P}$ represents the distribution induced by the altered model. A comparison of TV, SHD, and SID on these experiments is shown in Table~\ref{tbl:altered-models}. Two properties are apparent:
\begin{itemize}[noitemsep,topsep=0pt,parsep=0pt,partopsep=0pt]
    \item Model over-specification is not ignorable. For small datasets, such as that from the JDK domain, over-specified models have zero SID but significant TV values due to loss of statistical efficiency.
    \item Penalizing over-specification and under-specification with equal cost, as in SHD, is inconsistent with interventional distribution quality. In these domains, model over-specification has 2-5 times less distributional impact than under-specification as measured by total variation.
\end{itemize}

\subsection{REACTION TO STRENGTH OF DEPENDENCE}

\begin{figure}[t]
\centering
\begin{subfigure}{\linewidth}
    \includegraphics[width=\linewidth]{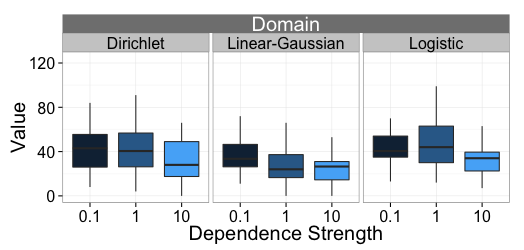}
    \caption{Structural intervention distance}
\end{subfigure}
\begin{subfigure}{\linewidth}
\includegraphics[width=\linewidth]{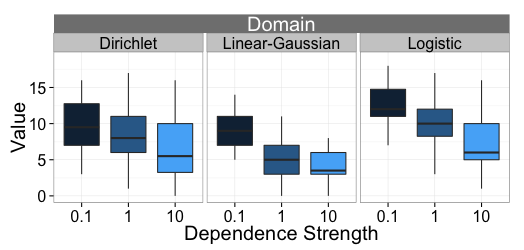}
\caption{Structural Hamming distance}
\end{subfigure}
\begin{subfigure}{\linewidth}
\includegraphics[width=\linewidth]{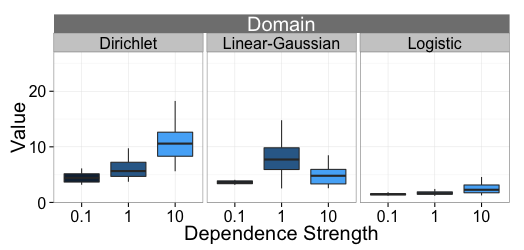}
\caption{Total variation distance}
\end{subfigure}
\caption{Performance measures with respect to strength of dependence. Each box contains 30 data points, with each data point representing performance on a CPDAG output by either PC, MMHC, or GES.}
\label{fig:strength-of-dependence}
\end{figure}

As already noted, an advantage of TV is that it weights inferred causal dependencies based on the strength of dependence.  However, a reasonable question is whether SID and SHD might still serve as reasonable proxies for TV as the strength of dependence varies.  We found that they do not.  Specifically, the presence of weak dependencies tends to increase structural measures (because learning weak dependencies is more difficult) but tends to decrease TV (because missing a weak dependence is less important).  To examine this effect, we again used experiments with synthetic data.
While many researchers use roughly similar parametric forms when generating data using the linear-Gaussian, logistic, and Dirichlet strategies, there is no accepted standard for the strength of dependence in each setting. For linear-Gaussian and logistic systems, strength of dependence can be controlled by adjusting the sampling distribution for the edge and vertex weights. In the case of the Dirichlet strategy, smaller $S$ values yield stronger dependencies (more skewed CPTs).

We generated variants of each synthetic domain with dependencies 10 times stronger or weaker than the most common value used in existing work. For each of these configurations, we generated 10 networks. We ran PC, MMHC, and GES on each of these networks, and recorded the mean SID, SHD, and TV of members of the resulting CPDAG. From Figure~\ref{fig:strength-of-dependence}, we see that the structural measures have an inverse relationship with TV as dependence strength is varied. As the strength of dependence decreases, the detectability of an effect is reduced, making structure learning more difficult. However, structural inaccuracies impact interventional distributions in accordance with strength of dependence---weak dependencies imply lower TV.

\section{DEVELOPING REALISTIC SYNTHETIC DATA}

\begin{figure}
\centering
\includegraphics[width=\linewidth]{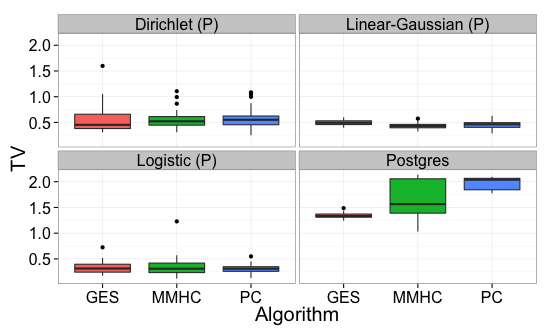}
\caption{Relative Performance on Postgres dataset and synthetic look-alike configurations.}
\label{fig:real-synthetic-compare-postgres}
\end{figure}

Now that we have established the value of TV for measuring the most important property of causal models (their ability to accurately estimate interventional distributions), we can deploy the measure to evaluate other properties of existing evaluation methods.  One key property is the inherent difficulty of learning causal models for various real and synthetic data sets.
We found that the real datasets were significantly more challenging than the synthetic datasets when measured with total variation.
One example is the Postgres dataset in which the best-performing model (GES) learns a much less accurate model than the worst performing algorithm on any synthetic domain (see Figure~\ref{fig:real-synthetic-compare-postgres}). The difference in means between each real domain and its synthetic counterparts is shown in Table~\ref{tbl:difference-in-means}. In all but one case (HTTP/Dirichlet), the real domains are more challenging.

\begin{table}
\centering
{ \small
\begin{tabular}{rrrr}
  \hline
 & Dirichlet & Linear-Gaussian & Logistic \\ 
  \hline
Postgres & 0.92 & 1.10 & 1.17 \\ 
  JDK & 2.20 & 2.51 & 3.05 \\ 
  HTTP & -0.31 & 0.45 & 0.54 \\ 
   \hline
\end{tabular}
}
\caption{Differences in means of total variation for all pairs of real domains and synthetic counterparts. All differences are strongly significant using Tukey's test.}
\label{tbl:difference-in-means}
\end{table}

We sought to characterize what alterations could be made to synthetic data generation techniques to reach the same level of difficulty as the real domains.
In the unbiased models for the three real datasets, it is common for treatment nodes to causally affect three or four outcomes. This stands in constrast to typical synthetic structures, for which nodes have one parent and one child in expectation.
To characterize the effect of this property on learned causal structures, we adjusted the synthetic data generation techniques to include additional edges between treatments and outcomes, and also varied the strength of dependence. 

Focusing on the largest dataset, Postgres, we generated synthetic configurations with treatments varying in out-degree from 3 to 5 and dependence strength varying from 1 to 5. For the linear-Gaussian and logistic techniques, all settings were significantly less challenging.
While the linear-Gaussian and logistic techniques can encode only a specific (generalized) linear form of dependence, the Dirichlet model can encode arbitrary multinomial CPTs. For the Dirichlet technique, some settings of the dependence strength and sparsity parameters yield datasets which are at least as challenging as Postgres (Table~\ref{tbl:sparsity-sweep-dirichlet}). This suggests that there are three key properties necessary for realistic synthetic evaluation: relatively dense connectivity, strong dependencies, and complex functional relationships.

\begin{table}
\setlength{\extrarowheight}{1pt}
\centering
{ \small
\begin{tabular}{ | r r | rrr | }
\hline
& & \multicolumn{3}{  c |}{Dependence Strength} \\
 & & 1 & 3 & 5 \\ 
 \hline
\multirow{3}{*}{\rotatebox[origin=c]{90}{Degree}} & 3 & 0.73 & 0.37 & - \\ 
  & 4 & 0.58 & -  & - \\ 
  & 5 & -  & -  & -0.93 \\ 
   \hline
\end{tabular}
}
\caption{Mean difference in pairwise TV between between Postgres and synthetic Dirichlet models with varying parameters. Positive numbers indicate that the Postgres configuration was more difficult than the Dirichlet counterpart. Hyphens indicate that no significant difference was present.}
\label{tbl:sparsity-sweep-dirichlet}
\end{table}

\section{CONCLUSIONS}

In this work, we provided empirical demonstrations that structural distance measures do not correspond to the quality of interventional distributions. Structural Hamming distance and structural intervention distance penalize for model over-specification in a way that is inconsistent with the parametric quality of estimated causal effects. Structural distances disagree with total variation distance as dependence strength varies, and can lead to different conclusions about relative algorithmic performance on a commonly used synthetic datasets.
Through a simple theoretical argument, we have shown that total variation distance captures a wider variety of modeling errors than structural intervention distance, and is more closely related to applications of causal discovery methods.

The synthetic datasets we studied are typically less challenging than the real datasets gathered from computational systems---suggesting that commonly employed synthetic evaluations have been unrealistically simple. We found that increasing network density, strength of dependence, and generating data with complex conditional models can yield synthetic models that are as challenging as real datasets.

\bibliographystyle{named}
\bibliography{causal-eval}

\end{document}